\documentclass{article}

\usepackage[T1]{fontenc} 
\usepackage[utf8]{inputenc} 

\usepackage{amssymb}
\usepackage{graphicx}
\usepackage{epstopdf}
\usepackage{color}
\usepackage{multirow}
\usepackage{hhline}
\usepackage{tikz}
\usepackage{nccmath}

\usepackage{soul}
\usepackage[skip=10pt]{subcaption}
\usepackage{ismir,cite,amsmath,url}

\title{Learning Disentangled Representations of\\Timbre and Pitch for Musical Instrument Sounds Using\\Gaussian Mixture Variational Autoencoders}

\usepackage{booktabs}

\multauthor
{Yin-Jyun Luo$^{1,2}$ \hspace{1cm} Kat Agres$^{2,3}$ \hspace{1cm} Dorien Herremans$^{1,2}$} { 
 $^1$ Singapore University of Technology and Design\\
$^2$ Institute of High Performance Computing, A*STAR, Singapore\\
$^3$ Yong Siew Toh Conservatory of Music, National University of Singapore\\
{\tt\small yinjyun\_luo@mymail.sutd.edu.sg,kat\_agres@ihpc.a\-star.edu.sg,dorien\_herremans@sutd.edu.sg}
}
\def\authorname{Yin-Jyun Luo, Kat Agres, Dorien Herremans}

\begin{document}

\maketitle
\begin{abstract}
In this paper, we learn disentangled representations of timbre and pitch for musical instrument sounds. We adapt a framework based on variational autoencoders with Gaussian mixture latent distributions. Specifically, we use two separate encoders to learn distinct latent spaces for timbre and pitch, which form Gaussian mixture components representing instrument identity and pitch, respectively. For reconstruction, latent variables of timbre and pitch are sampled from corresponding mixture components, and are concatenated as the input to a decoder. We show the model's efficacy using latent space visualization, and a quantitative analysis indicates the discriminability of these spaces, even with a limited number of instrument labels for training. The model allows for controllable synthesis of selected instrument sounds by sampling from the latent spaces. To evaluate this, we trained instrument and pitch classifiers using original labeled data. These classifiers achieve high F-scores when tested on our synthesized sounds, which verifies the model’s performance of controllable realistic timbre/pitch synthesis. Our model also enables timbre transfer between multiple instruments, with a single encoder-decoder architecture, which is evaluated by measuring the shift in the posterior of instrument classification. Our in-depth evaluation confirms the model's ability to successfully disentangle timbre and pitch.\footnote{Example audio files and code at \texttt{http://bit.ly/2Dbyt9j}}

\end{abstract}

\section{Introduction}\label{sec:introduction}

A disentangled feature representation is defined as having disjoint subsets of feature dimensions that are only sensitive to changes in corresponding factors of variation from observed data~\cite{kumar2017variational, bengio2013deep, ridgeway2016survey}. Deep generative models~\cite{hu2017unifying, kingma2013auto, goodfellow2014generative, oord2016wavenet} have been exploited to learn disentangled representations in different domains. In the visual domain, studies are focused on learning independent representations for data generative factors such as identity and azimuth~\cite{chen2016infogan, kulkarni2015deep, higgins2017beta}. In natural language generation, efforts have been made to generate texts with controlled sentiment~\cite{hu2017toward, zhou2018emotional, fu2018style}. Also in the speech domain, we have witnessed successful attempts in controllable speech synthesis by disentangling factors such as speaker identity, speed of speech, emotion, and noise level~\cite{hsu2017unsupervised, wang2018style, hsu2018hierarchical}. There has been relatively little research on learning disentangled representations for music. In this paper, we disentangle the pitch and timbre of musical instrument sound recordings.

Pitch and timbre are essential properties of musical sounds. Given that one pitch can be played with different instruments, we assume they can be separated. From the perspective of music analysis, disentangled representations of pitch and timbre can be regarded as timbre- and pitch-invariant features which could be exploited for downstream tasks~\cite{muller2011signal, muller2010towards}. From the synthesis point of view, disentangled representations enable the generation of musical notes with identical pitches (timbres) and different timbres (pitches). Recently, Hung \textit{et al.} presented the first attempt to learn disentangled representations of pitch and timbre for synthesized music by using frame-level instrument and pitch labels based on encoder-decoder networks~\cite{hung2018learning}. Even though the authors managed to change instrumentation to some extent without affecting pitch structure, the approach was restrictive, as it worked with MIDI-synthesized audio and relied on clean frame-level labels, which are scarce to find. Disentangled representations allow for several applications, including music style transfer. Brunner \textit{et al.} proposed a model based on variational autoencoders (VAEs)~\cite{kingma2013auto} to generate music with controllable attributes~\cite{brunner2018midi}. While genre was factorized by an auxiliary classifier, other musical properties were entangled. Besides the aforementioned models based on MIDI, research on audio has focused on translating between different domains of instrumentation~\cite{engel2017neural, mor2018universal, bitton2018modulated, huang2018timbretron}. None of them, however, has addressed learning disentangled latent variables of both pitch and timbre.

This research distinguishes itself from others by disentangling instrument sounds into distinct sets of latent variables (i.e., pitch and timbre), with a framework based on Gaussian Mixture VAEs (GMVAEs). We model the generative process of an isolated musical note by independently sampling pitch and timbre (instrument) categorical variables. Note that the two factors are actually dependent in a sense that range of pitch is instrument-dependent, however, we verify the model's capability to disentangle them under this simplified assumption of independence.
Conditioned on these categorical variables, Gaussian-distributed latent variables are then sampled that characterize variation in the sampled pitch and instrument, respectively. Finally, the data are generated conditioned on the two latent variables. We favor the proposed framework over vanilla VAEs~\cite{esling2018bridging, esling2018generativetimbre} for its more flexible latent distribution compared to a standard Gaussian. In addition, it allows for unsupervised or semi-supervised clustering, which can learn interpretable mixture components and corresponding Gaussian parameters. More importantly, such a framework facilitates the applications in this research: controllable synthesis of instrument sounds, and many-to-many transfer of instrument timbres. Our proposed framework differs from previous studies on timbre transfer, in that we achieve transfer between \emph{multiple instruments} without training a domain-specific decoder for each instrument (e.g.~\cite{mor2018universal}), and we infer both the pitch and timbre latent variable without requiring categorical conditions of source pitch and instrument as in~\cite{bitton2018modulated}. We evaluate our model by visualizing both the latent space and the synthesized spectrograms, and explore the classification F-scores of classifiers trained in an end-to-end fashion. The results confirm the model's ability to learn disentangled pitch and timbre representations. The rest of the paper is organized as follows: in Section~\ref{sec:GMVAE}, we discuss the proposed framework, and Section~\ref{sec:setup} describes the dataset and experimental setup. Experiments and results are reported in Section~\ref{sec:experiment}. We conclude our work and provide future directions in Section~\ref{sec:conclusion}.

\begin{figure}[t]
\centering
 \includegraphics[width=0.9\columnwidth]{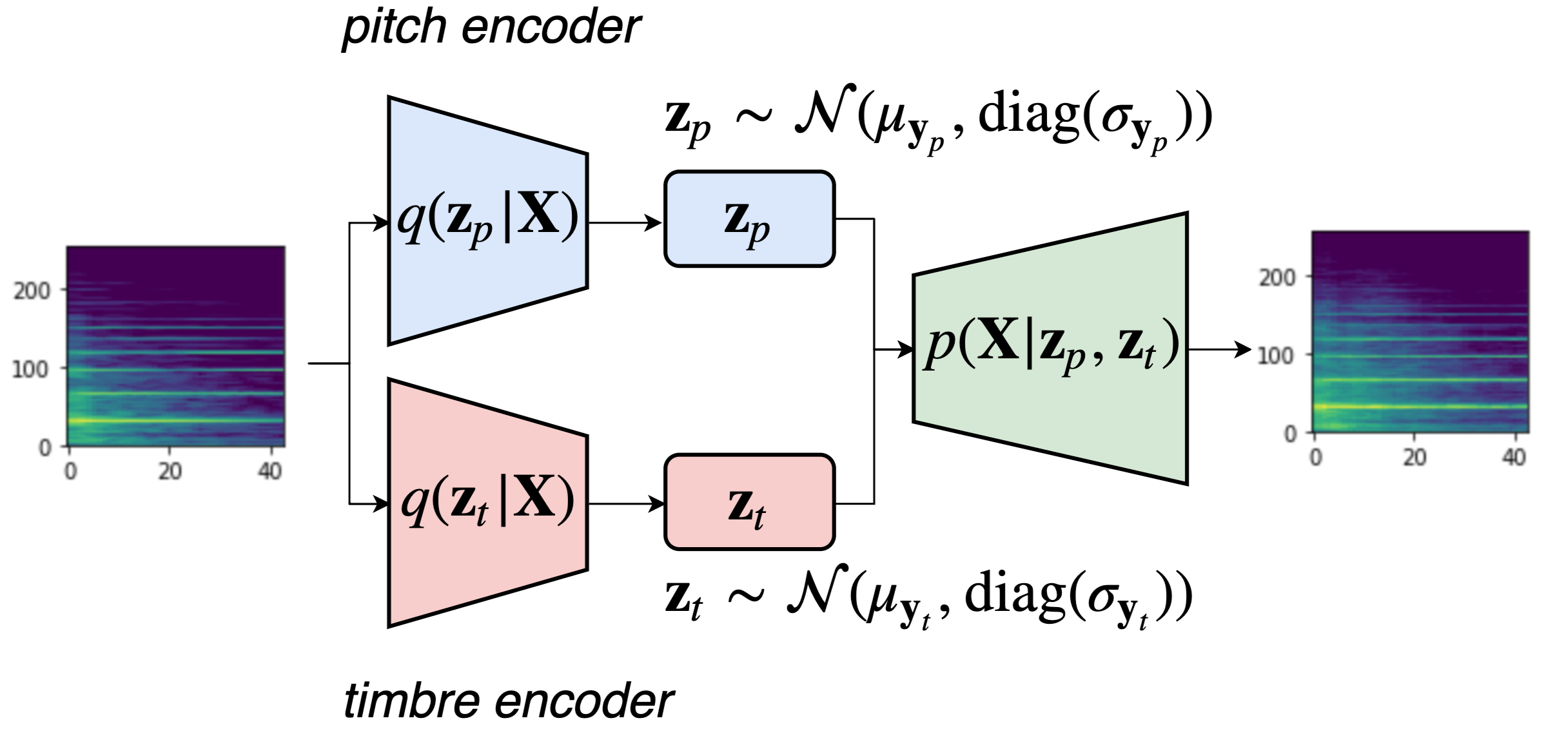}
 \caption{The proposed framework includes separate encoders for pitch and timbre, and a shared decoder.} 
 \label{fig:architecture}
\end{figure}

\section{Proposed Framework}\label{sec:GMVAE}
In this section, we briefly describe VAEs and GMVAEs, and elaborate on the proposed framework and architecture.

\subsection{Gaussian Mixture Variational Autoencoders}
VAEs~\cite{kingma2013auto} are unsupervised generative models that combine latent variable models and deep learning~\cite{goodfellow2016deep}. We denote the observed data and the latent variables respectively by $\mathbf{X}$ and $\mathbf{z}$. A graphical model, corresponding to $\mathbf{z}\rightarrow\mathbf{X}$, is trained by maximizing the lower bound of the log marginal likelihood $p(\mathbf{X})$. The intractable posterior $p(\mathbf{z} | \mathbf{X})$ is approximated by introducing a variational distribution $q(\mathbf{z} | \mathbf{X})$ parameterized with neural networks. In regular VAEs, a common choice for the prior distribution $p(\mathbf{z})$ is an isotropic Gaussian, which encourages each dimension of the latent variables to capture an independent factor of variation from the data, and results in a disentangled representation~\cite{higgins2017beta}. Such a unimodal prior, however, does not allow for multi-modal representations. GMVAEs~\cite{dilokthanakul2016deep, kingma2014semi, jiang2016variational} extend the prior to a mixture of Gaussians, and assume the observed data are generated by first determining the mode from which it was generated, which corresponds to learning a graphical model $\mathbf{y}\rightarrow\mathbf{z}\rightarrow\mathbf{X}$. This introduces a categorical variable $\mathbf{y}$, and $q(\mathbf{y}|\mathbf{X})$, which infers the classes of data. This enables semi-supervised learning~\cite{kingma2014semi} and unsupervised clustering~\cite{dilokthanakul2016deep, jiang2016variational} in deep generative models. In the speech domain, Hsu \textit{et al.} used two mixture distributions to separately model the supervised speaker and unsupervised utterance attributes, which allowed for extra flexibility in conditional speech generation~\cite{hsu2018hierarchical}. We build upon this idea to learn separate latent distributions to represent the pitch and timbre of musical instrument sounds. More importantly, to facilitate downstream creative applications such as controllable synthesis and instrument timbre transfer in music, we propose to model supervised pitch representations and semi-supervised timbre representations, with labels of pitch and instrument identity. As such, the mixture components in latent space of pitch and timbre can be clearly interpreted as the classes, i.e., pitch and instrument identity.
\subsection{Model Formulation}\label{subsec:model}
The latent variables of pitch and timbre for an isolated musical note $\mathbf{X}$ are denoted as ${\mathbf{z}_{p}}$ (\textit{pitch code}) and ${\mathbf{z}_{t}}$ (\textit{timbre code}), respectively. To represent Gaussian mixture latent distributions, two categorical variables are introduced: an \textit{M}-way categorical variable $\mathbf{y}_{p}$ for pitch, where \textit{M} is the number of recorded pitches in the dataset, and a \textit{K}-way categorical variable $\mathbf{y}_{t}$ for timbre, where \textit{K} is the number of instrument classes. We consider $\mathbf{y}_{p}$ to be observed (fully supervised), which assumes the availability of pitch labels during training, and is reasonable as we model isolated instrument sounds in this research. For $\mathbf{y}_{t}$, we investigate both unsupervised and semi-supervised learning, i.e., using varying numbers of instrument labels for training. It is shown in Section~\ref{sec:experiment} that our model can efficiently leverage the limited number of labels. Without loss of generality, we denote $\mathbf{y}_{t}$ as unobserved (unsupervised) as in~\cite{hsu2018hierarchical}. The joint probability of $\mathbf{X}$, $\mathbf{y}_t$, $\mathbf{z}_t$ and $\mathbf{z}_p$ is written as:
\begin{equation}\label{jointprob}
\begin{split}
    p(\mathbf{X}, \mathbf{y}_{t}, &\mathbf{z}_{t}, \mathbf{z}_{p} | \mathbf{y}_{p}) = \\
    & p(\mathbf{X} | \mathbf{z}_{p}, \mathbf{z}_{t}) p(\mathbf{z}_{p} | \mathbf{y}_{p}) p(\mathbf{z}_{t} | \mathbf{y}_{t}) p(\mathbf{y}_{t}),
\end{split}
\end{equation}
where $p(\mathbf{y}_{t})$ is uniform-distributed, i.e., we do not assume to know the instrument distribution in the dataset. Both the conditional distributions $p(\mathbf{z}_{p} | \mathbf{y}_{p}) = \mathcal{N}(\bm{\mu}_{\mathbf{y}_{p}}, \textrm{diag}(\bm{\sigma}_{\mathbf{y}_{p}}))$ and $p(\mathbf{z}_{t} | \mathbf{y}_{t}) = \mathcal{N}(\bm{\mu}_{\mathbf{y}_{t}}, \textrm{diag}(\bm{\sigma}_{\mathbf{y}_{t}}))$ are assumed to be diagonal-covariance Gaussians with learnable means and constant variances. This amounts to both the marginal prior $p(\mathbf{z}_{p})$ and $p(\mathbf{z}_{t})$ being Gaussian mixture models (GMMs) with diagonal covariances. Ideally, each mixture component in the former (\textit{pitch space}) uniquely represents the pitch of $\mathbf{X}$, while that in the latter (\textit{timbre space}) is interpreted as the instrument identity. As we will see in Section~\ref{sec:tsne}, however, moderate supervision is essential to learn a timbre space that groups instruments perfectly. For creative applications such as the synthesis and timbre transfer of instrument sounds, the proposed model has numerous merits: 1) the learnt representations are not restricted to be unimodal, which offers a more discriminative timbre space than regular VAEs (Section~\ref{sec:tsne} and~\ref{sec:disentangle}); 2) direct and intuitive sampling from pitch and timbre space allows for consistent and controllable synthesis of instrument sounds, attributed to the fact that Gaussian parameters of each interpretable mixture component are readily available after training (Section~\ref{sec:control}); and 3) simple arithmetic manipulations between means of mixture components facilitate many-to-many transfer between instrument timbres (Section~\ref{sec:transfer}). For the training objective, we closely follow the derivation in~\cite{hsu2018hierarchical} and train the model by maximizing the evidence lower bound (ELBO) as follows:
\begin{equation}\label{elbo}
\begin{split}
    \mathcal{L}(p,q;\mathbf{X}, \mathbf{y}_{p}) &= \mathbb{E}_{
    q(\mathbf{z}_{p} | \mathbf{X}) 
    q(\mathbf{z}_{t} | \mathbf{X})}[\log p(\mathbf{X}|\mathbf{z}_{p},\mathbf{z}_{t})] \\
    &-D_{KL}(q(\mathbf{z}_{p} | \mathbf{X}) || p(\mathbf{z}_{p} | \mathbf{y}_{p}))\\
    &-\mathbb{E}_{q(\mathbf{y}_{t}|\mathbf{X})}[
    D_{KL}(q(\mathbf{z}_{t}|\mathbf{X}) | p(\mathbf{z}_{t} | \mathbf{y}_{t})] \\
    &-D_{KL}(q(\mathbf{y}_{t} | \mathbf{X}) || p(\mathbf{y}_{t})),
\end{split}
\end{equation}
where $p(\mathbf{X} | \mathbf{z}_{p}, \mathbf{z}_{t})$, $q(\mathbf{z}_{p} | \mathbf{X})$, and $q(\mathbf{z}_{t}| \mathbf{X})$ are parameterized with neural networks, referred to as the \textit{decoder}, \textit{pitch encoder}\footnote{A common alternative is conditioning the model with categorical pitch labels such that one does not have to train a pitch encoder \protect{~\cite{engel2017neural, bitton2018modulated}}. It, however, requires the pitch of the inputs to be known a priori to performing tasks such as timbre transfer \protect{\cite{bitton2018modulated}}, and also prohibits the model from extracting pitch features for downstream tasks. By training this extra encoder, we also demonstrate how one can extend the model to possibly learn multiple interpretable latent variables.}, and \textit{timbre encoder}, respectively. Instead of using another neural network, we approximate $q(\mathbf{y}_{t} | \mathbf{X})$ by $\mathbb{E}_{q(\mathbf{z}_{t}|\mathbf{X})}[p(\mathbf{y}_{t}|\mathbf{z}_{t})]$. Readers interested in detailed derivation are referred to Appendix A in~\cite{hsu2018hierarchical}.

\subsection{Architecture}
Our model is composed of a shared decoder and separate encoders for pitch and timbre, as illustrated in \figref{fig:architecture}.  Specifically, we reshape the $T$-by-$F$ spectrogram to have number of channels $C = F$, each of which is a $T$-by-$1$ vector, where $T$ and $F$ refer to time and frequency. Each encoder contains two one-dimensional convolutional layers, each with 512 filters of shape $3 \times 1$, and a fully connected layer with 512 units. A Gaussian parametric layer follows and outputs two $L-$dimensional vectors which represent mean and log variance. $\mathbf{z}_{p}$ and $\mathbf{z}_{t}$ are sampled from the Gaussian layer with the reparameterization trick~\cite{kingma2013auto}, which enables stochastic gradient descent, and are then concatenated for the decoder to reconstruct the input. The architecture of the decoder is symmetric to the encoder. Batch normalization followed by the activation function \texttt{relu} are used for every layer except for the Gaussian and the output layer. We use the activation function \texttt{tanh} for the output layer as we normalize the data within $[-1, 1]$.

\begin{figure*}[ht]

\begin{subfigure}[b]{\textwidth}
\centering
 \includegraphics[width=0.7\textwidth, trim={0 1.20cm 0 0}, clip]{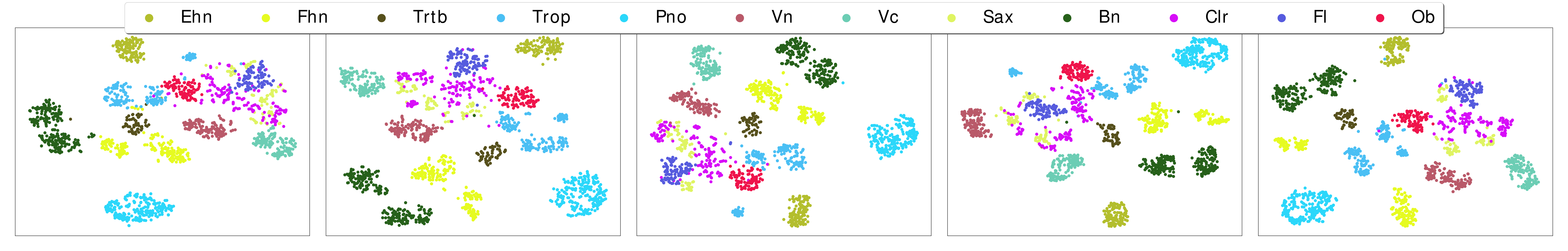}
 \label{fig:vae_tsne}
 \end{subfigure}
\begin{subfigure}[b]{\textwidth}
\centering
 \includegraphics[width=0.7\textwidth, trim={0 0 0 1cm}, clip]{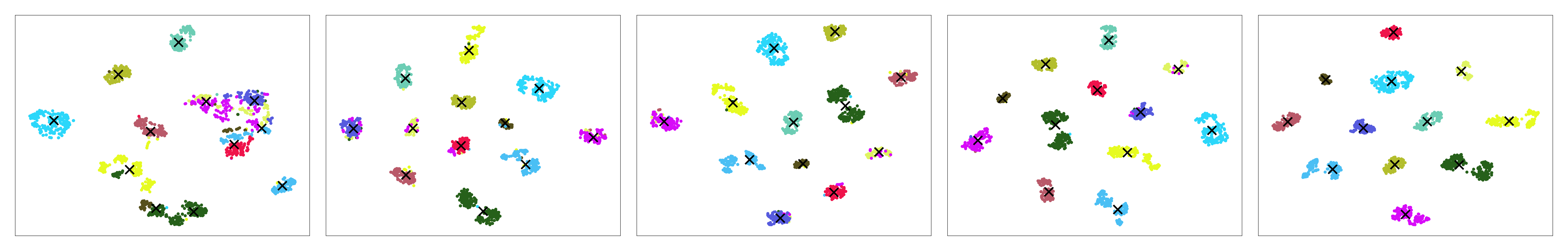}
 \label{fig:gmvae_tsne}
\end{subfigure}
\vspace{-17pt}
\caption{Timbre space visualization of M\textsubscript{VAEs} (top) and M\textsubscript{GMVAEs} (bottom). From left to right: models trained with 0, 25, 50, 75, or 100\% of instrument labels, respectively.}
\label{fig:tsne}
\end{figure*}

\begin{table*}[ht]
\centering
\small

\begin{tabular}{c|ccccc|ccccc} 
\toprule
      & \multicolumn{5}{c|}{Instrument Classification}  & \multicolumn{5}{c}{Pitch Classification} \\ 
\midrule
$N$ (\%) & CNN &  \multicolumn{2}{c}{M\textsubscript{VAE}} &  \multicolumn{2}{c|}{M\textsubscript{GMVAE}} & CNN &  \multicolumn{2}{c}{M\textsubscript{VAE}} &  \multicolumn{2}{c}{M\textsubscript{GMVAE}}\\
      &     & $\mathbf{z}_{t}$  & $\mathbf{z}_{p}$ & $\mathbf{z}_{t}$ & $\mathbf{z}_{p}$ &   &  $\mathbf{z}_{t}$ & $\mathbf{z}_{p}$ & $\mathbf{z}_{t}$ & $\mathbf{z}_{p}$ \\ 
\midrule
\midrule
0      & -     & 0.960 & 0.163 & 0.937 & 0.175 & -     & 0.112 & 0.966 & 0.146 & 0.960 \\
25     & 0.920 & 0.960 & 0.192 & 0.971 & 0.180 & -     & 0.169 & 0.966 & 0.084  & 0.977 \\
50     & 0.983 & 0.971 & 0.169 & 0.988 & 0.186 & -     & 0.158 & 0.977 & 0.079  & 0.977 \\
75     & 1.000 & 0.971 & 0.169 & 1.000 & 0.163 & -     & 0.079  & 0.971 & 0.045  & 0.977 \\
100    & 1.000 & 0.937 & 0.158 & 1.000 & 0.197 & 0.983 & 0.039  & 0.983 & 0.028  & 0.966 \\

\bottomrule
\end{tabular}
\caption{The F-scores of instrument and pitch prediction by linear classifiers and CNNs. $N$ (\%) refers to the percentage of instrument labels used to train the models. Columns $\mathbf{z}_t$ and $\mathbf{z}_p$, respectively, refer to the F-scores obtained using the learned timbre and pitch code to train the down-stream linear classifier.}
\label{tab:classification}
\end{table*}

\section{Experimental Setup}\label{sec:setup}
In this section, we describe the experimental setup, including details of the dataset, input representations, and model configurations.

\subsection{Dataset}\label{sec:dataset}
Inspired by Esling \textit{et al.}~\cite{esling2018bridging}, we use a subset of Studio-On-Line (SOL)~\cite{sol1999}, a database of instrument note recordings.\footnote{Access to the dataset was requested from \protect{\cite{esling2018bridging}}.} The dataset contains 12 instruments, i.e, piano (\texttt{Pno}, 246), violin (\texttt{Vn}, 138), cello (\texttt{Vc}, 147), English horn (\texttt{Ehn}, 128), French horn (\texttt{Fhn}, 214), tenor trombone (\texttt{Trtb}, 63), trumpet (\texttt{Trop}, 194), saxophone (\texttt{Sax}, 99), bassoon (\texttt{Bn}, 251), clarinet (\texttt{Clr}, 180), flute (\texttt{Fl}, 118) and oboe (\texttt{Ob}, 107). There are 1,885 samples in total. All recordings are resampled to 22,050Hz, and only the first 500ms segment ($T=43$) of each recording is considered. We extract Mel-spectrograms with 256 filterbanks ($F=256$), derived from the power magnitude spectrum of the short-time Fourier transform (STFT). To compute STFT, we use a Hann window with window size of 92ms and hop size of 11ms. As a result, the input representation is a 43-by-256 Mel-spectrogram. The dataset is split into a training (90\%) and validation set (10\%), each containing the same distribution of instruments. The magnitude of the Mel-spectrogram is scaled logarithmically, and the minimum and maximum values in the training set are used for normalizing the magnitude within $[-1, 1]$ in a corpus-wide fashion to preserve differences in dynamics.

\subsection{Hyperparameters}
In order to train both the GMMs in pitch and timbre space, we initialize the means of mixture components using Xavier initialization~\cite{glorot2010understanding}. We set constant standard deviations, rather than trainable ones, for pitch and timbre space. For pitch space, $\bm{\sigma}_{\mathbf{y}_{p}} = \mathbf{e^{-2}}$ for all mixture components, which is relatively small, as each mixture component represents a pitch, and we do not expect a large variance over recordings that play the same pitch. For timbre space, we let $\bm{\sigma}_{\mathbf{y}_{t}} = \mathbf{e^{0}}$ for all mixture components, which captures the timbre variation of each mixture component, i.e., instrument identity. The dimensionality of the latent space is $L = 16$, and the numbers of mixture components are $M = 82$ and $K = 12$, equivalent to the numbers of classes of pitch and instrument, respectively. For all experiments, a batch size of 128 is used, model parameters are initialized with Xavier initialization and are trained using the Adam optimizer~\cite{kingma2014adam} with a learning rate of $10^{-4}$.

In addition to the proposed model (\textit{M\textsubscript{GMVAE}}), we consider a baseline (\textit{M\textsubscript{VAE}}) that substitutes the timbre space with an isotropic Gaussian as in regular VAEs. Training such a model amounts to optimizing \eqnref{elbo} with the last two terms replaced with $D_{KL}(q(\mathbf{z}_{t} | \mathbf{X}) || p(\mathbf{z}_{t}))$, where $p(\mathbf{z}_{t}) = \mathcal{N}(\boldsymbol{0}, \boldsymbol{I})$. The experimental results in Section~\ref{sec:tsne} and Section~\ref{sec:disentangle} show that M\textsubscript{GMVAE} learns a more discriminative and disentangled timbre space than M\textsubscript{VAE}.

\subsection{Semi-Supervised Learning}
We exploit a moderate number of instrument labels to learn a timbre space in which the clusters clearly represent instrument identity. Similar to Kingma \textit{et al.}~\cite{kingma2014semi}, in the semi-supervised training for M\textsubscript{GMVAE}, we \textit{guide} the inference of instrument labels $q(\mathbf{y}_{t} | \mathbf{X})$ by leveraging limited amounts of supervision. This is done by adding an additional loss term which measures the cross entropy between the inferred and true instrument labels. Because we do not infer $\mathbf{y}_t$ in M\textsubscript{VAE}, we use $\mathbf{z}_t$ to train an auxiliary classifier to predict $\mathbf{y}_t$. It has two 128-unit fully-connected layers, and is jointly optimized with M\textsubscript{VAE}. We consider varying numbers of instrument labels $N$ = 0 (unsupervised), 25, 50, 75, and 100\% (fully supervised) of the total number. We randomly sample and let the label distribution match the distribution of instruments.

\section{Experiments and Results}\label{sec:experiment}
The experiments and the results are presented in this section. We first visualize the timbre space, and quantitatively evaluate the disentangled representations. We then demonstrate the applications of controllable synthesis and many-to-many timbre transfer. Finally, we identify the particular latent dimension that is sensitive to the distribution of the spectral centroid, which allows for finer timbre controls.
\subsection{Visualization}\label{sec:tsne}
\figref{fig:tsne} visualizes the timbre space using t-distributed stochastic neighbor embedding (t-SNE)~\cite{maaten2008visualizing}, a technique that projects vectors from high- to low-dimensional space. We first observe that M\textsubscript{GMVAE} learns a Gaussian-mixture distributed timbre space, with means of mixture components marked as crosses in the figure. Second, attributed to the pitch encoder which addresses pitch variations, both M\textsubscript{VAE} and M\textsubscript{GMVAE} are able to form clusters of instrument identity even without being trained with instrument labels (the leftmost column). We observe that the wind family (e.g., saxophone, clarinet and flute) forms an ambiguous cluster. Such an ambiguity remains in the M\textsubscript{VAE} even with increased $N$, while it is less present in the M\textsubscript{GMVAE} latent space, due to the multi-modal prior distribution. As we will confirm in Section~\ref{sec:disentangle}, M\textsubscript{GMVAE} outperforms M\textsubscript{VAE} in learning a more discriminative and disentangled timbre space. Note that in M\textsubscript{GMVAE}, $p(\mathbf{y}_{t})$ is assumed to be uniformly distributed over 12 classes of instruments, i.e., mixture components are equally weighted. As a result, instruments with larger within-class variances (e.g., bassoon and trumpet) are assigned to more than one cluster when $N = 0$. In future work we aim to improve the performance of the unsupervised clustering of instruments. 
\subsection{Pitch and Instrument Disentanglement}\label{sec:disentangle}

A disentangled pitch (timbre) representation should be discriminative for pitch (instrument identity), and at the same time non-informative of instrument identity (pitch). Therefore, we evaluate $\mathbf{z}_p$ and $\mathbf{z}_t$ by means of classification. We train linear classifiers to map $\mathbf{z}_p$ and $\mathbf{z}_t$ to predict both pitch and instrument labels with one fully connected layer. For comparison, we train an end-to-end convolutional neural network (CNN), whose architecture is the same as the encoder and is a strong baseline, to map the original input Mel-spectrograms to either pitch or instrument labels.

\tabref{tab:classification} shows the results. The CNN achieves high F-scores on both instrument and pitch classification; note that $N$ is the supervisory percentage of the total number of \textit{instrument} labels, and we always use all pitch labels to train the models, which is reasonable as we model isolated notes in this work. In instrument classification, using $\mathbf{z}_t$ as the feature representations outperforms $\mathbf{z}_p$ by a large margin, as expected. Specifically, in both models, the $\mathbf{z}_t$ learned with unsupervised learning ($N$ = 0)
is already discriminative enough to predict instruments with linear classifiers. While the F-score of M\textsubscript{GMVAE} improves with increased $N$, that of M\textsubscript{VAE} does not. Moreover, the linear classifier trained with $\mathbf{z}_t$ outperforms the CNN when $N < 75$. The timbre space of M\textsubscript{GMVAE} displays the most discriminative power among the models.
We attribute the F-scores of instrument classification attained by $\mathbf{z}_p$ to the fact that the piano covers all possible pitches in the dataset, while other instruments account for a smaller pitch range. As a result, $\mathbf{z}_p$ of notes that were only recorded by piano are correctly classified. Future work can be done to decorrelate particular pitches and instruments by data augmentation and adversarial training as in~\cite{hsu2018disentangling}. In pitch classification, $\mathbf{z}_p$ outperforms $\mathbf{z}_t$ as expected, and both models achieve comparable results. More importantly, M\textsubscript{GMVAE} performs better than M\textsubscript{VAE} in terms of disentanglement, as $\mathbf{z}_t$ results in lower F-scores when predicting pitch with increased N.
\begin{figure}[!t]
\centering
 \includegraphics[width=\columnwidth]{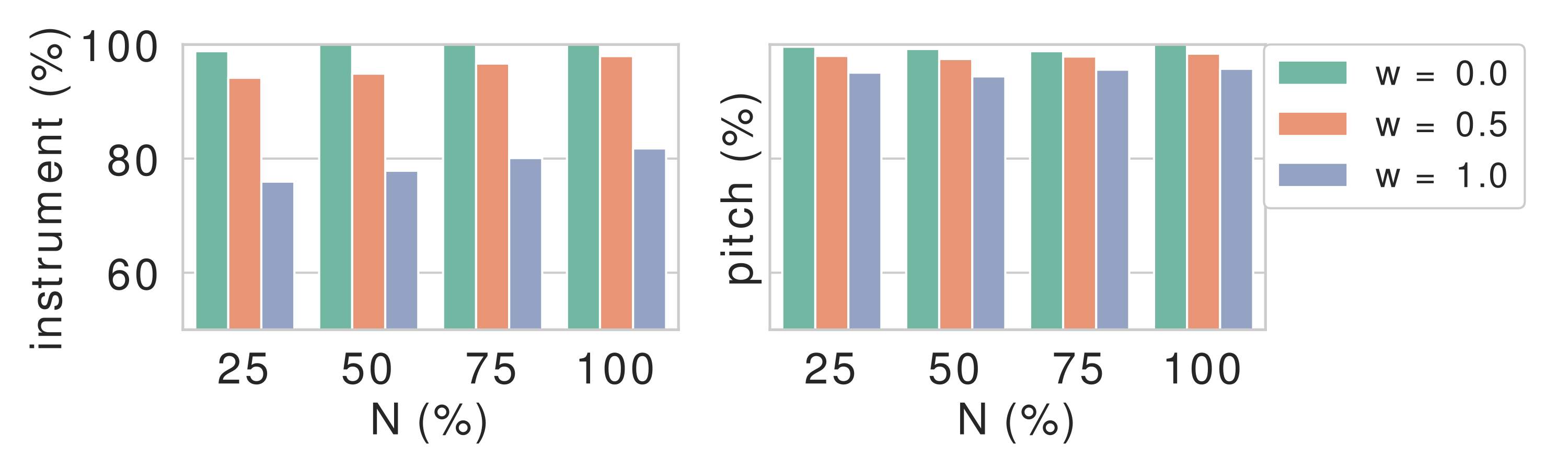}
\caption{The F-scores for predicting instrument (left) and pitch (right) labels from the synthesized spectrograms.}
\label{fig:controllable}
\end{figure}
\subsection{Controllable Synthesis of Instrument Sounds}\label{sec:control}
As shown in \figref{fig:tsne}, M\textsubscript{GMVAE} learns a timbre space $p(\mathbf{z}_t)$, whose mixture components are clearly interpreted as instrument identity when trained with moderate supervision. Meanwhile, mixture components in $p(\mathbf{z}_p)$ represent pitch. As Gaussian parameters are readily available after training, we can achieve controllable sound synthesis by sampling $p(\mathbf{z}|\mathbf{y})$.
To synthesize the target pitch $y_m$ and instrument $y_k$, we first sample $\mathbf{z}_p \sim \mathcal{N}(\bm{\mu}_{y_m}, w\cdot \textrm{diag}(\bm{\sigma}_{y_m}))$ and $\mathbf{z}_t \sim \mathcal{N}(\bm{\mu}_{y_k}, w\cdot \textrm{diag}(\bm{\sigma}_{y_k}))$, where the multiplier $w \in \{0, 0.5, 1.0\}$ serves to examine the effect of sampling latent variables that deviate from the modes. The decoder then synthesizes the Mel-spectrogram by consuming $[\mathbf{z}_t, \mathbf{z}_p]$. For evaluation, the CNNs (trained on the original dataset) are used to test whether the synthesized spectrograms are still recognized as belonging to the desired instrument and pitch. High F-scores therefore indicate high controllability of the model in sound synthesis. We use the sound samples in the validation set as the targets to synthesize, and repeat the sampling 30 times for each target. 

The F-scores for pitch and instrument classification are reported in \figref{fig:controllable}. We first note that increasing $w$ degrades classification performance. This is expected, as a sample which is synthesized using a latent variable far from its corresponding mean of mixture component deviates more from the intended instrument or pitch distribution. Moreover, the fact that the CNN was trained on the original samples while tested on the synthesized ones also contributes to the inferior performance. Second, increasing $N$ improves instrument classification performance. Finally, the high F-scores across all $N$'s when $w \in \{0, 0.5\}$ indicate accurate and consistent synthesis of instrument sounds with intended pitches and instruments, even with a timbre space trained using a limited number of instrument labels. This implies that M\textsubscript{GMVAE} efficiently exploits the instrument labels, and learns a discriminative mixture distribution of timbre, which is consistent with the visualization in \figref{fig:tsne} (bottom row, $N\ge25$).
We do not explore the timbre space resulting from unsupervised learning ($N = 0$) in this experiment, as the instrument identity of each mixture component is not directly available. We can, however, infer the instrument identity of each mixture component by sampling and synthesis, and expect reasonably good performance for controllable synthesis if the clustering of instruments shown in the bottom left of \figref{fig:tsne} is improved. This will be explored in future work.

\begin{figure}[!t]
\centering

 \includegraphics[width=\columnwidth, trim={0 0 10cm 0}, clip]{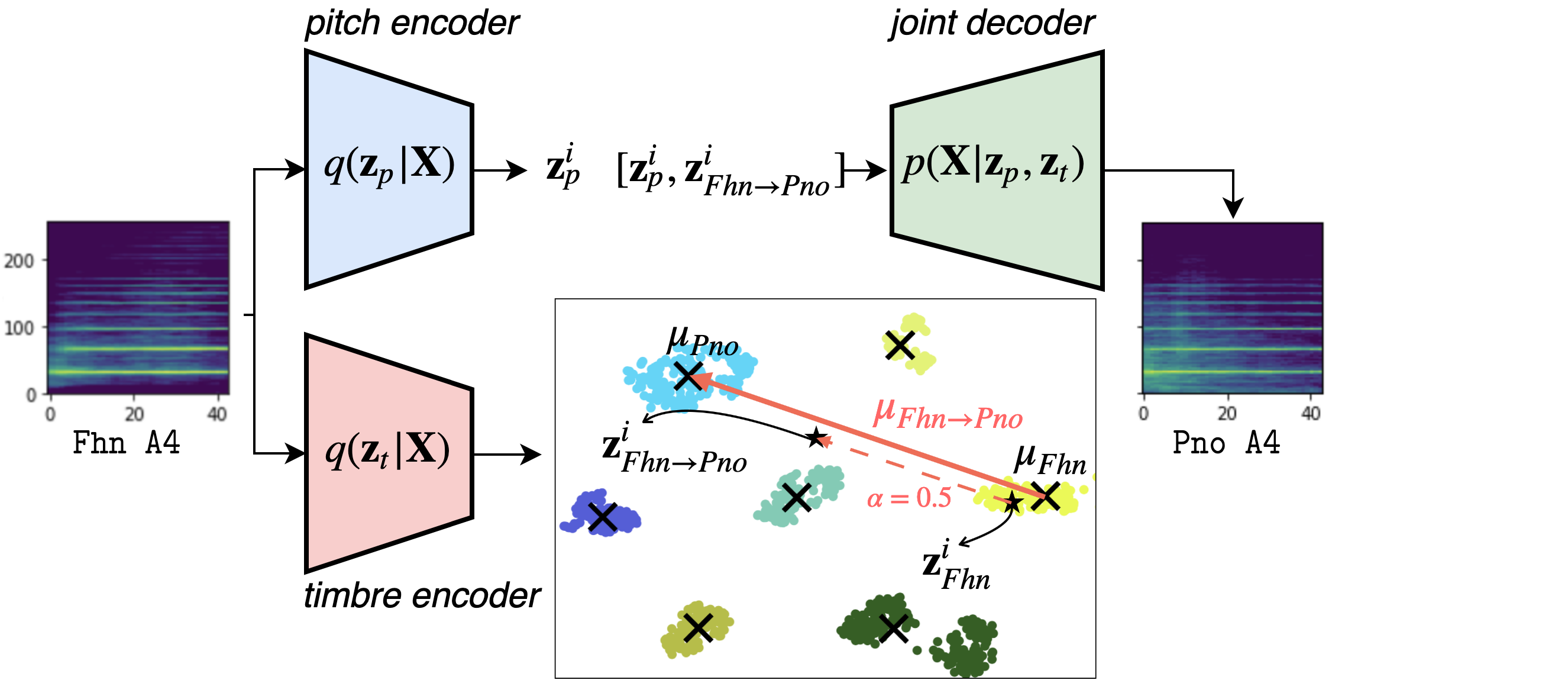}
 \caption{Many-to-many timbre transfer. The $i$th sample of the \texttt{Fhn} is transferred to the \texttt{Pno}, with vector arithmetic in the (partially shown) timbre space.}
 \label{fig:timbretransfer}
\end{figure}

\subsection{Many-to-Many Transfer of Timbre}\label{sec:transfer}
In this experiment, we demonstrate many-to-many transfer of instrument timbre. In Mor \textit{et al.}, a domain-specific decoder was trained for each target~\cite{mor2018universal}. To achieve timbre transfer with a single encoder-decoder architecture, Bitton \textit{et al.} proposed to use a conditional layer~\cite{perez2018film} which takes both instrument and pitch labels as inputs~\cite{bitton2018modulated}. On the other hand, our model infers $\mathbf{z}_t$ and $\mathbf{z}_p$, and only uses a single joint decoder. As illustrated in \figref{fig:timbretransfer}, timbre transfer is achieved by decoding $[\mathbf{z}_{transfer}, \mathbf{z}_p]$, i.e., transferring timbre while keeping pitch unchanged, where $\mathbf{z}_{transfer} = \mathbf{z}_{source} + \alpha\bm{\mu}_{source \rightarrow target}$, $\bm{\mu}_{source \rightarrow target} = \bm{\mu}_{target} - \bm{\mu}_{source}$, and $\alpha \in [0, 1]$. Once again, we rely on the trained CNNs in \tabref{tab:classification} for evaluation. More specifically, we examine the posterior shift in instrument prediction of the CNN, before and after transferring from source to target instruments with $\alpha = \{0, 0.25, 0.5, 0.75, 1.0\}$. For simplicity, the most frequent instruments (i.e., French horn, piano, cello, and bassoon) of the four families are selected as the representatives, and we perform timbre transfer using the samples in the validation set as the source. For example, consider \texttt{Fhn} as the source and \texttt{Pno} as target, as shown in \figref{fig:timbretransfer}. We modify the timbre code as $\mathbf{z}^{i}_{Fhn \rightarrow Pno} = \mathbf{z}^{i}_{Fhn} + \alpha\bm{\mu}_{Fhn \rightarrow Pno}$, where $\mathbf{z}^{i}_{Fhn}$ is the timbre code of the $i$th \texttt{Fhn} sample, and $i = \{1, 2, \ldots, \textrm{N}\textsubscript{Fhn}\}$. We decode as described earlier and report the averaged posterior (over N\textsubscript{Fhn}) of instrument prediction of the CNN.

\begin{figure}[!t]
\centering
\includegraphics[width=\columnwidth, trim={0 0 0 1.1cm}]{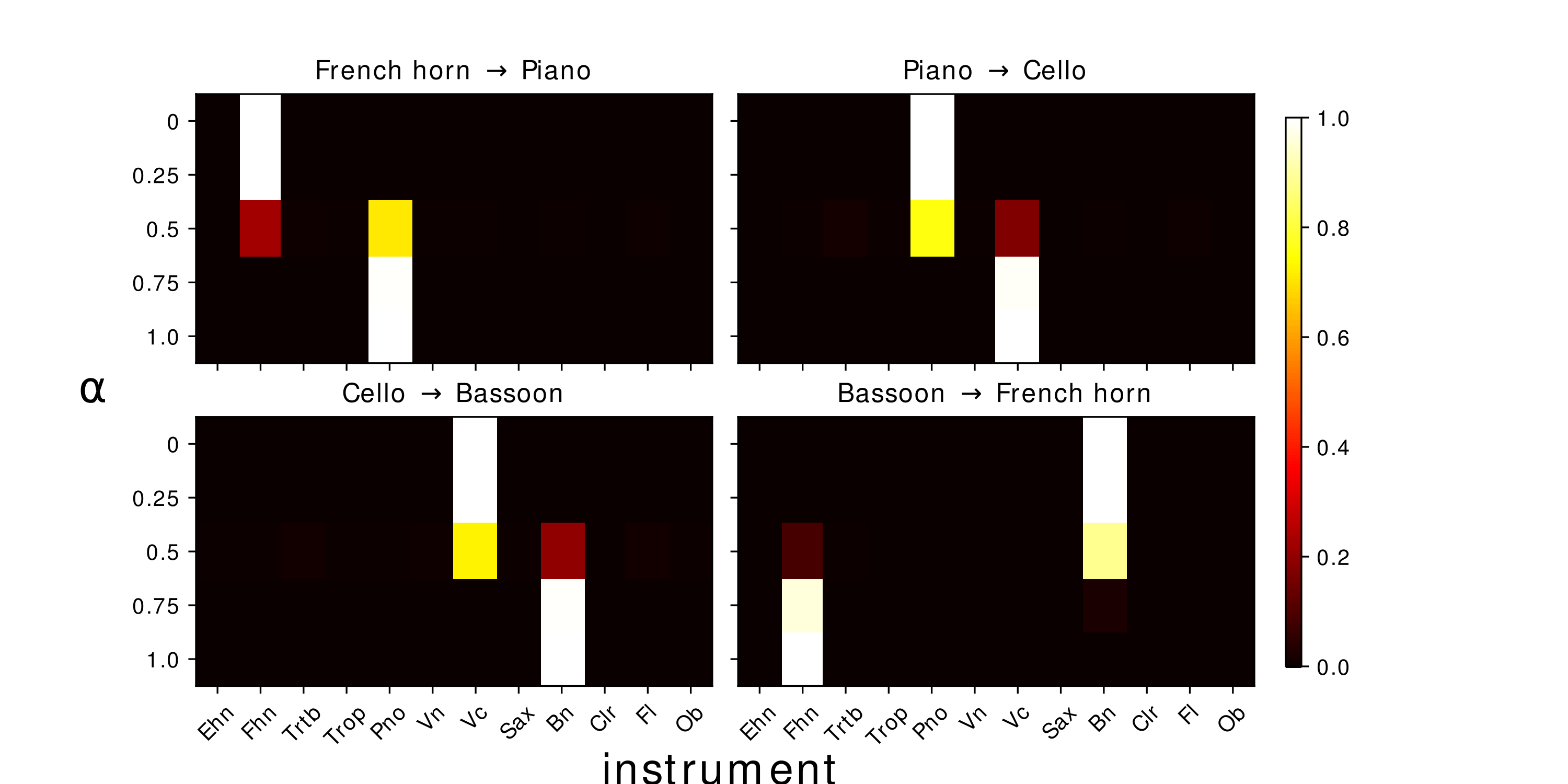}
\caption{The averaged posterior (color) shift in instrument prediction of the CNN, caused by timbre transfer.}

\label{fig:transfer}
\end{figure}


\begin{figure}[!b]
\centering
 \includegraphics[width=.90\columnwidth, trim={0 0 0 0.3cm}, clip]{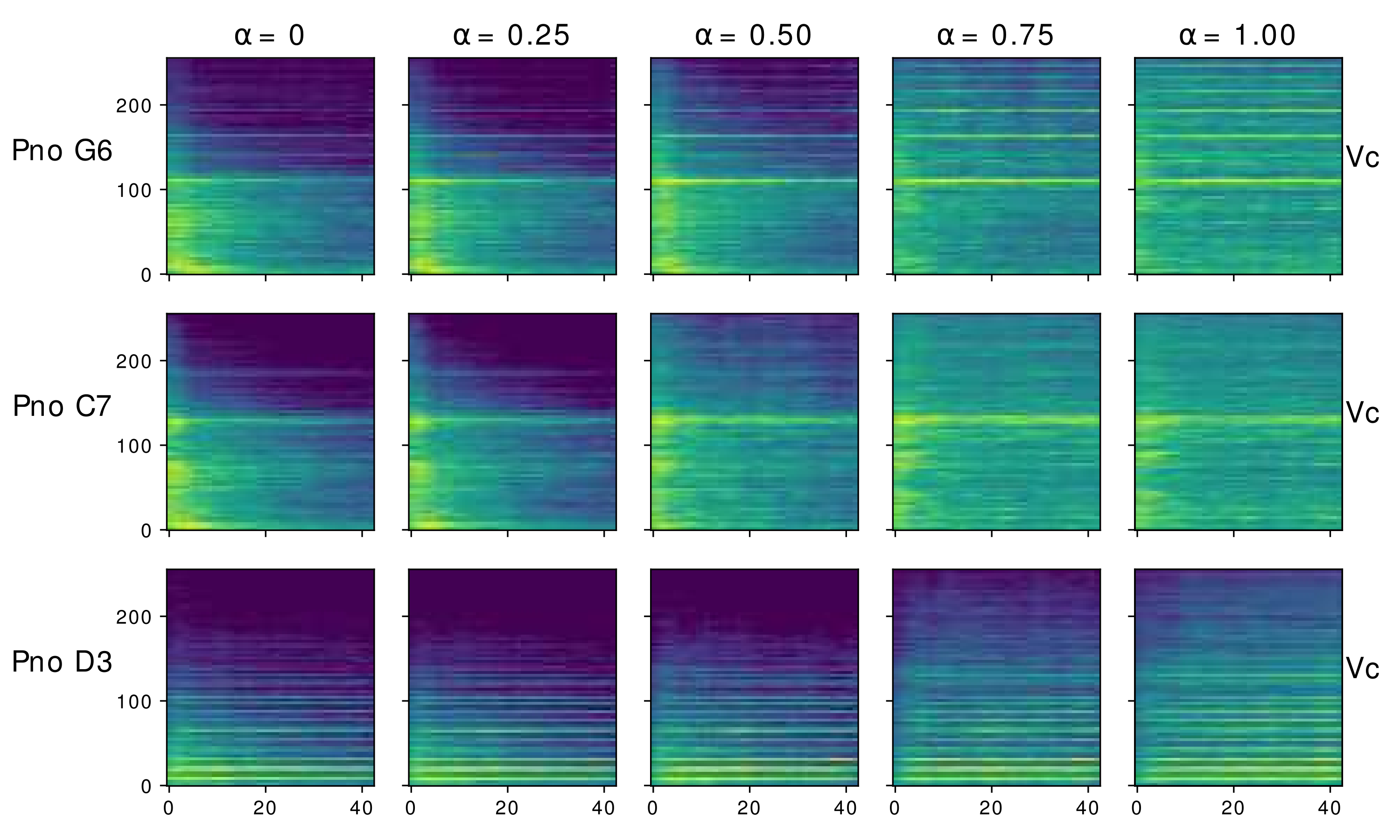}
 \caption{Examples of timbre transfer $Pno\rightarrow Vc$. The top two rows are tones outside of the cello range.}
 \label{fig:spec_transfer}
\end{figure}

\begin{figure}[!t]
\centering  
 \includegraphics[width=\columnwidth]{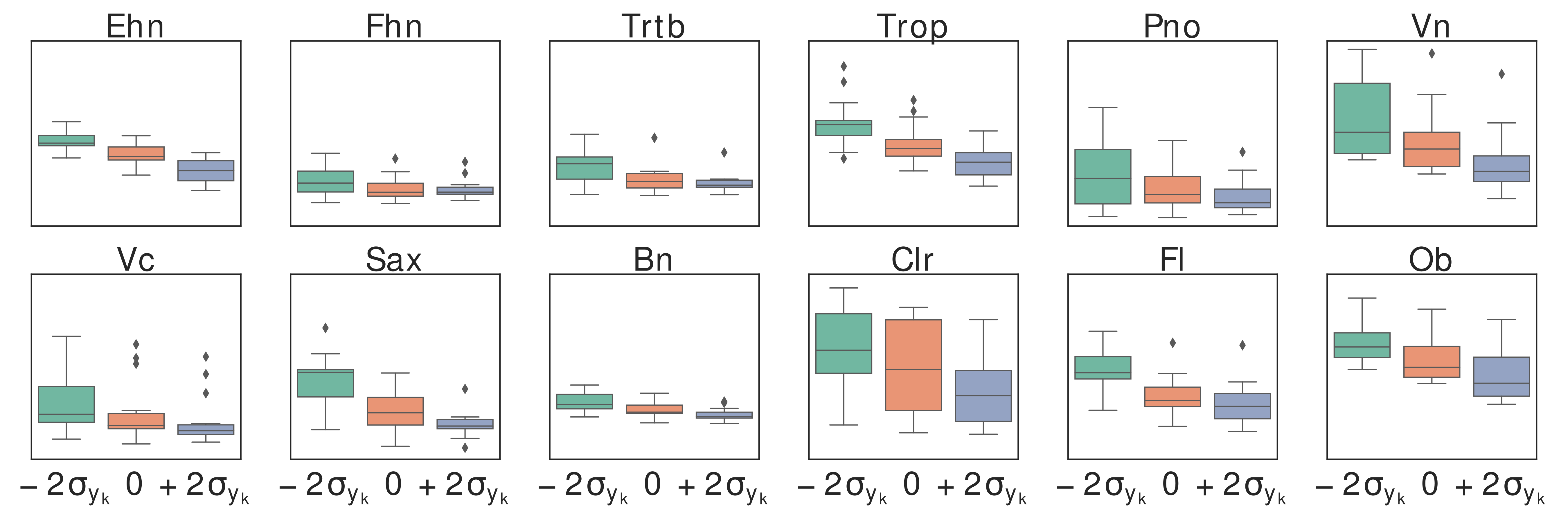}
 
 \caption{Spectral centroid values in response to $z^{13}_{t}$.}
 \label{fig:spec_centroid}
\end{figure}

For simplicity, in \figref{fig:transfer}, we report the results of the source-target pairs $Fhn\rightarrow Pno$, $Pno\rightarrow Vc$, $Vc\rightarrow Bn$ and $Bn\rightarrow Fhn$. Each subfigure refers to a source-target pair, and represents the averaged posterior shift of instrument classification of the CNN, with varying $\alpha$. For all pairs, the biggest posterior shift (hence the prediction change) happens when $\alpha = 0.5$. This also applies to the rest of the possible instrument pairs not shown in the figure. Meanwhile, by using pitch classification, we examine if the pitches are the same before and after timbre transfer, and we use the original pitch labels as ground-truths. We find that, except in the case where the source is piano, all source-target pairs attain a perfect F-score in terms of pitch. This confirms the ability of the model to successfully perform many-to-many timbre transfer. A special case arises when piano is the source. The F-scores before transfer, after transfer to French horn, to cello, and to bassoon, are 0.958, 0.750, 0.791, and 0.791, respectively. As described earlier in Section~\ref{sec:disentangle}, lower F-scores can be attributed to the fact that the range of piano is much larger than that of the target instruments, or the classifier fails to predict the synthesized samples that have unseen combinations of pitch and instrument. The other possible reason is the model falls short of generalization. Nevertheless, this only happens in some cases when the source is piano; as demonstrated in \figref{fig:spec_transfer}, the model is able to transfer \texttt{Pno G6} to cello (the first row), which is an example of generalizing to an out-of-range pitch for the target instrument. In the first and third row, the high-frequency components appear with increased $\alpha$, and the energy distributes over the segment without decay. The model, however, falls short in generalizing to the higher pitch, i.e., \texttt{Pno C7} (the second row), where the energy remains focused at the onset, and high-frequency components are smeared. In the future, we could improve the model generalizability by performing data augmentation and adversarial training as in~\cite{hsu2018disentangling}.

\subsection{Spectral Centroid Disentanglement}
\begin{figure}[!h]
\vspace{-5pt}
\centering  
 \includegraphics[width=.9\columnwidth, trim={0 1cm 0 0.8cm}, clip]{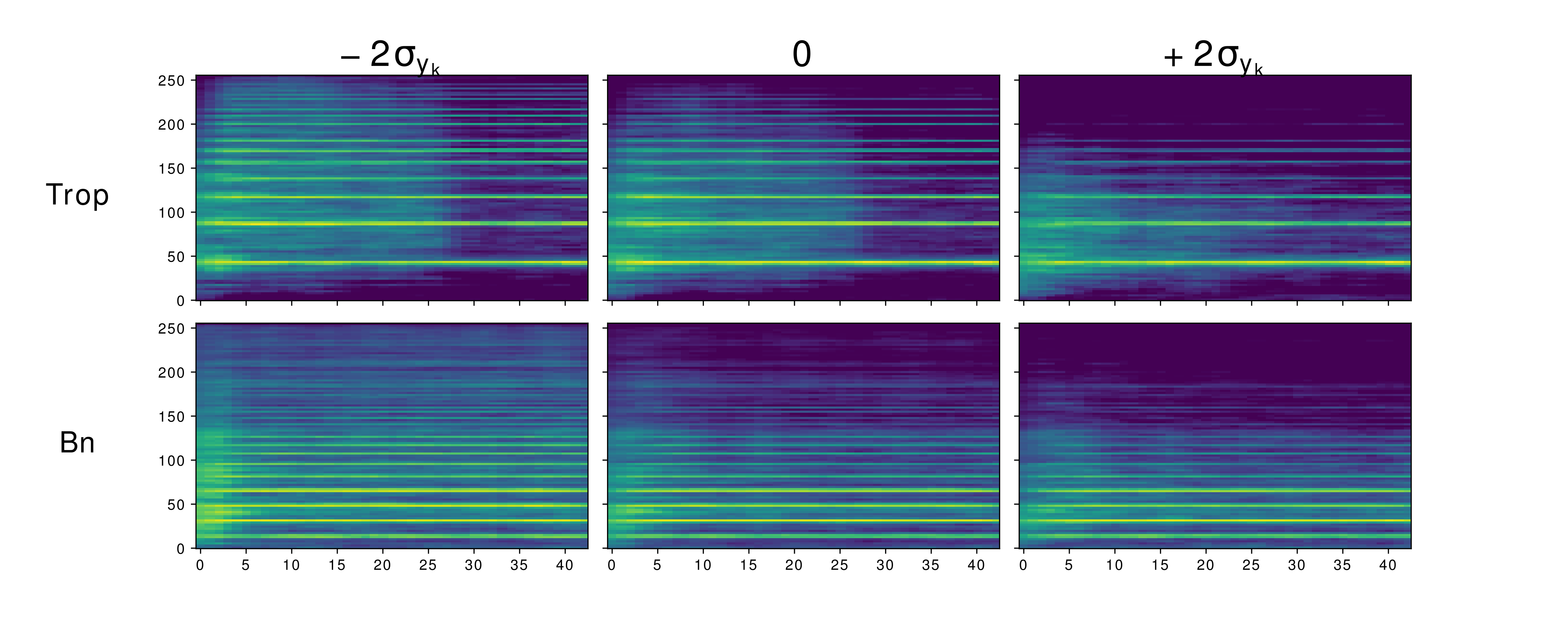}
 \caption{Latent dimension traverse of $z^{13}_{t}$.}
 \vspace{-12pt}
 \label{fig:spec_centroid_spec}
\end{figure}

A diagonal-covariance Gaussian prior encourages the model to learn disentangled latent dimensions~\cite{higgins2017beta}. This applies to all mixture components in our model. In particular, we identify a latent dimension that correlates with the spectral centroid. 
we modify the 13th dimension of $\mathbf{z}_t$, $z^{13}_{t}$, of each sound sample in the validation set by $\pm 2\sigma_{y_{k}}$, where $\sigma_{y_{k}} = e^{0}$ for all instruments, and then synthesize the spectrograms, for which we then calculate the spectral centroid. \figref{fig:spec_centroid} shows the distributions of the spectral centroid before and after the modifications. The two-tailed t-test indicates significant differences ($p<0.05$) between $-2\sigma_{y_k}$ and $+2\sigma_{y_k}$ for all instruments. As demonstrated in \figref{fig:spec_centroid_spec}, we observe that increased $z^{13}_{t}$ reduces the energy of high-frequency components and results in lower spectral centroid values. In future research, we will further investigate disentangling specific acoustic features for finer control of sound synthesis beyond pitch and instrument.

\section{Conclusions and Future Work}\label{sec:conclusion}
We have proposed a framework based on GMVAEs to learn disentangled timbre and pitch representations for musical instrument sounds, which is verified by our experimental setup. We demonstrate its applicability in controllable sound synthesis and many-to-many timbre transfer. In future work, we plan to conduct listening tests for a more comprehensive evaluation of the applications, and further disentangle both low- (e.g., acoustic features) and high-level (e.g., playing techniques) sound attributes, enabling finer control of synthesized timbres. By using supervised and unsupervised learning in a deep generative model, the framework can be easily adapted to learn interpretable mixtures such as singer identity, music style, emotion, etc., which facilitates music representation learning and creative applications.

\section{Acknowledgments}
We would like to thank the anonymous reviewers for their constructive reviews. This work is supported by a Singapore International Graduate Award (SINGA) provided by the Agency for Science, Technology and Research (A*STAR), under reference number SING-2018-01-1270.

\end{document}